\pdfoutput=1
\documentclass[conference, 10pt]{IEEEtran}
\IEEEoverridecommandlockouts

\usepackage{color}
\usepackage{xcolor}
\usepackage{theorem}
\usepackage{times,amsmath,epsfig}
\usepackage{amssymb}
\usepackage{subfig}
\usepackage{cite}
\usepackage{url}
\usepackage{moresize}

\usepackage{graphicx} 
\usepackage{hyperref}
\usepackage{bm}
\usepackage{amsmath}

\input{mysymbol.sty}

\newcommand{\bphi}{\bm{\phi}}

\newcommand{\phimap}{\hbphi_{\mathrm{MAP}}}
\newcommand{\Thetaml}{\hbTheta_{\mathrm{GL}}}
\newtheorem{problem}{Problem}

\title{Enhancing Graphical Lasso: A Robust Scheme \\
for Non-Stationary Mean Data}

\author{S. Rey$^{\dagger}$, E. Curbelo$^{\ddagger}$,  L. Martino$^\star$, F. Llorente$^\diamond$, A. G. Marques$^{\dagger}$ \\
{\footnotesize$^{\dagger}$ Dept. of Signal Theory and Communications, Universidad Rey Juan Carlos, Madrid, Spain} \\
{\footnotesize$^{\ddagger}$ Dept. of Statistics, Universidad Carlos III de Madrid, Madrid, Spain.} \\
{\footnotesize$^\star$ Dept. of Economics and Business, Universit{\`a} degli Studi di Catania, Catania, Italy.}\\
{\footnotesize$^\diamond$ Computing and Data Science
Directorate, Brookhaven National Laboratory, Upton, USA.}\\
\thanks{This work was supported in part by the Spanish AEI Grants PID2022-136887NB-I00, PID2023-149457OB-I00, and the Community of Madrid (CAM) and Rey Juan Carlos University (URJC) via the Young Researchers R\&D Project F1180-AdvGSP4BIO, and Madrid ELLIS Unit, by the "PIAno di inCEntivi per la RIcerca di Ateneo 2026” (UPB 28722052159) and by
PIACERI Starting Grant BA-GRAPH (UPB 28722052144) of the University of Catania.}}

\begin{document}

\maketitle
\begin{abstract}
This work addresses the problem of graph learning from data following a Gaussian Graphical Model (GGM) with a time-varying mean.  
Graphical Lasso (GL), the standard method for estimating sparse precision matrices, assumes that the observed data follows a zero-mean Gaussian distribution.  
However, this assumption is often violated in real-world scenarios where the mean evolves over time due to external influences, trends, or regime shifts.  
When the mean is not properly accounted for, applying GL directly can lead to estimating a biased precision matrix, hence hindering the graph learning task.
To overcome this limitation, we propose \emph{Graphical Lasso with Adaptive Targeted Adaptive Importance Sampling (GL-ATAIS)}, an iterative method that jointly estimates the time-varying mean and the precision matrix.
Our approach integrates Bayesian inference with frequentist estimation, leveraging importance sampling to obtain an estimate of the mean while using a regularized maximum likelihood estimator to infer the precision matrix.  
By iteratively refining both estimates, GL-ATAIS mitigates the bias introduced by time-varying means, leading to more accurate graph recovery. 
Our numerical evaluation demonstrates the impact of properly accounting for time-dependent means and highlights the advantages of GL-ATAIS over standard GL in recovering the true graph structure.  
\end{abstract}
%

\section{Introduction}
A fundamental problem in data analysis is the inference of pairwise interactions between variables, which can be naturally framed as learning the topology of a graph.  
Graphs serve as powerful tools for modeling structured data in diverse fields such as neuroscience, genomics, and finance~\cite{belilovsky2016testing,smith2011network,stegle2015computational}.  
Among the many approaches for inferring graph structure from nodal observations~\cite{kalofolias2016learn,segarra2017network,mateos2019connecting}, Graphical Lasso (GL) stands out as a prominent and widely used method~\cite{lauritzen1996graphical,friedman2008sparse,egilmez2017graph,buciulea2025polynomial}.  
GL operates under the assumption that the data follows a Gaussian Graphical Model (GGM), where the graph structure is encoded in the precision matrix, i.e., the inverse of the covariance matrix.  
By leveraging this assumption, GL provides a regularized maximum likelihood estimator for inferring the underlying graph.  

Over the years, numerous extensions of GL have been developed to enhance its robustness and flexibility.
These include adaptations for scenarios with partial observability~\cite{chandrasekaran2012latent,meng2014learning,buciulea2022learning}, more sophisticated structural priors to encode advanced graph information~\cite{kumar2019structured,navarro2022joint,rey2023enhanced,buciulea2025polynomial}, and methods for jointly estimating multiple graphs~\cite{jacob2009group,danaher2014joint,rey2022joint}.
Despite these advancements, a key underlying assumption remains: the observed data follows a \emph{zero-mean} multivariate Gaussian distribution.  
While in some cases the sample mean can be removed as a preprocessing step, this approach is inadequate in many real-world settings.  
For instance, time series data frequently exhibit time-dependent mean structures due to external influences, trends, or regime shifts.  
If the mean is not properly estimated before applying GL, the inferred precision matrix may be biased, leading to incorrect graph topology estimation.  
Thus, accounting for temporal variability in the mean is crucial for ensuring reliable graph learning with GL.  

Several works have explored graph learning in dynamic environments.  
The works in~\cite{hallac2017network,greenewald2017time,cai2018capturing,yang2020estimating} consider scenarios where the observations follow a GGM with a time-evolving precision matrix, which implies that the graph changes over time.
Building on this premise, later approaches have developed efficient online algorithms for learning dynamic graph structures from streaming data, hence adapting to changes in graph connectivity~\cite{yang2010online,natali2022learning,rey2025online}.  
Despite their importance, these methods generally assume that the mean of the observed data is zero, which limits their applicability in settings where the mean is time-dependent.
More related to our work is~\cite{curbelo2025adaptive}, which focuses on estimating the time-dependent mean of a GGM, with the covariance matrix obtained as a byproduct.  
However, this approach does not explicitly address the problem of graph topology inference, and the only available information about the precision matrix comes from the inverse of the sample covariance, which is known to be a poor estimate in high-dimensional regimes.  

This work addresses the aforementioned limitations of GL by considering the problem of graph learning when observations are drawn from a GGM with a time-dependent mean.  
Accurately estimating the mean is essential to prevent bias in the learned precision matrix.  
To this end, we propose a novel approach, termed Graphical Lasso with Adaptive Targeted Adaptive Importance Sampling (GL-ATAIS), which integrates Bayesian inference with frequentist estimation to overcome the limitations of GL in the presence of time-dependent means.
GL-ATAIS is an alternating optimization algorithm where we first leverage importance sampling to estimate the time-dependent mean, and then harness this estimate to center the observations and apply GL to learn the precision matrix.
By iteratively refining both the mean and the precision matrix estimates, GL-ATAIS provides a principled approach for recovering the true graph structure under time-varying mean conditions.

\section{Fundamentals of Gaussian Graphical Models}
In this section, we provide the necessary definitions and background for our problem of interest.

\vspace{2mm}
\noindent\textbf{Graphs and graph signals}.
Let $\ccalG := (\ccalV, \ccalE)$ be an undirected and weighted graph, where $\ccalV$ denotes the set of nodes and $\ccalE \subseteq \ccalV \times \ccalV$ represents the set of edges.  
The graph consists of $|\ccalV| = N$ nodes, and for any pair $(i,j) \in \ccalV \times  \ccalV$, an edge $(i,j) \in \ccalE$ exists if and only if nodes $i$ and $j$ are connected.  
The connectivity of $\ccalG$ is encoded in the weighted adjacency matrix $\bbA \in \reals^{N \times N}$, where $A_{ij} \neq 0$ if and only if $(i,j) \in \ccalE$.
Together with the graph $\ccalG$, we consider signals defined on its node set $\ccalV$.  
Formally, a \emph{graph signal} is a function from the node set to the real field, $x: \ccalV \to \reals$, or equivalently, an $N$-dimensional vector $\bbx \in \reals^N$, where $x_i$ denotes the signal value at node $i$.   

\vspace{2mm}
\noindent\textbf{Gaussian Graphical Models (GGMs)}.
GGMs describe the conditional dependence structure of multivariate Gaussian distributions as a graph.  
Consider a random graph signal $\bbx \in \reals^N$ drawn from a Gaussian distribution with mean $\bbmu$ and covariance matrix $\bbSigma \succ 0$, i.e., $\bbx \sim \ccalN(\bbmu, \bbSigma)$.  
The inverse covariance matrix $\bbTheta\! = \! \bbSigma^{-1}$, known as the \emph{precision matrix}, fully characterizes the conditional dependence structure of $\bbx$.  
Specifically, for any distinct pair $(i,j)\! \in \!\ccalV\!\times\!\! \ccalV$, the variables $x_i$ and $x_j$ are conditionally independent if and only if $\Theta_{ij} = 0$~\cite{lauritzen1996graphical, koller2009probabilistic}.  
This Markovian property naturally induces a graph $\ccalG$, where the nonzero off-diagonal entries of $\bbTheta$ define the weighted edges representing conditional dependencies between variables.

\vspace{2mm}
\noindent\textbf{Graphical Lasso (GL)}.
Let the matrix $\bbX = [\bbx_1, \dots, \bbx_R] \in \reals^{N \times R}$ collect $R$ independent samples drawn from $\ccalN(\bbzero, \bbTheta^{-1})$.  
A widely used method to estimate the graph structure encoded in $\bbTheta$ from $\bbX$ is GL~\cite{friedman2008sparse}, which solves the following convex optimization problem
\begin{align}
    \hbTheta &= \argmin_{\bbTheta \succeq 0} \frac{1}{2R} \sum_{r=1}^R \bbx_r^{\top}\bbTheta \bbx_r - \frac{1}{2}\log\det (\bbTheta) + \lambda \| \bbTheta \|_1, \nonumber \\ 
   &= \argmin_{\bbTheta \succeq 0} \;\; \frac{1}{2}\tr(\hbSigma_{\texttt{emp}}\bbTheta) - \frac{1}{2}\log\det (\bbTheta) + \lambda \| \bbTheta \|_1. \label{eq:GL_problem}
\end{align}
Here, $\hbSigma_{\texttt{emp}} = \frac{1}{R} \bbX \bbX^\top$ is the empirical covariance matrix, the constrain $\bbTheta \succeq 0$ represents the set of positive semidefinite matrices, and $\| \bbTheta \|_1$ is the elementwise $\ell_1$-norm of $\bbTheta$ excluding its diagonal entries to promote sparsity.
Indeed, the objective can be interpreted as a negative log-likelihood function, which we use in the next section to define the Bayesian model.

The GL formulation in~\eqref{eq:GL_problem} provides a regularized maximum likelihood estimator of $\bbTheta$, relying on the assumption that the mean of the distribution is zero, i.e., $\bbmu = \mbE [\bbx] = \bbzero$.  
When $\bbmu \neq \bbzero$, the standard approach is to use the sample mean estimate $\hbmu = \frac{1}{R}\sum_{r=1}^R \bbx_r$ to center the data before applying GL.  
However, when $\bbmu$ varies over time, $\hbmu$ becomes a poor estimate, leading to inaccurate recovery of the graph topology encoded in $\bbTheta$.  
To overcome this limitation, we next propose an iterative procedure where we first estimate the time-varying mean $\bbmu$ and then infer $\bbTheta$ using the debiased observations.  
This strategy ensures a more accurate estimation of the underlying graph structure in scenarios where the mean exhibits temporal variations.  

\section{GGMs with Time-varying mean}
This work focuses on the general setting where the observations $\bbx_r$ follow a random time series modeled by a GGM with a time-varying mean, i.e., $\bbx_r \sim \ccalN(\bbmu_r, \bbTheta^{-1})$, with $r$ denoting a discrete time index.
Furthermore, we assume that the mean $\bbmu_r$ can be accurately represented by a (possibly non-linear) parametric function $\bbf_r(\bbphi)$, where $\bbphi \in \reals^{M}$ denotes the unknown parameters, and the subscript $r$ emphasizes the time dependency.  
Intuitively, this is equivalent to the following signal model
\begin{equation}\label{eq:time_dep_ggm}
    \bbx_r = \bbf_r (\bbphi) + \bbv_r,
\end{equation}  
where $\bbv_r \sim \ccalN(\bbzero, \bbTheta^{-1})$ represents zero-mean fluctuations modeled by a standard GGM. We note that $\bbf_r$ is a vector valued-funtion (meaning that the time-varying function associated with each node can be different). 
With this notation at hand, we formally state our graph learning problem below.

\begin{problem}\label{p:problem_statement}
    Let the matrix $\bbX \in \reals^{N \times R}$ collect $R$ observations of a time series following~\eqref{eq:time_dep_ggm}.  
    Our goal is to estimate the graph topology encoded in the precision matrix $\bbTheta \in \reals^{N \times N}$, assuming that the function $\bbf_r(\bbphi)$ with unknown parameters $\bbphi \in \reals^M$, can be evaluated.  
\end{problem}

While our primary interest lies in learning the graph structure, accurately estimating the parameters $\bbphi$ is crucial, as it enables an improved estimation of the mean of the underlying process.
Notably, we do not assume prior knowledge of the parametric function $\bbf_r(\bbphi)$, nor do we impose any specific structure, such as convexity or differentiability, thereby allowing for a broad range of possible mean functions.
With these considerations in mind, we now introduce an iterative approach designed to address this challenge and solve Problem~\ref{p:problem_statement}.


\subsection{GL with MAP estimation of time-varying mean}

The cornerstone of our method is accurately estimating the mean of the GGM, which amounts to estimating the parameters of the function $\bbf_r(\bbphi)$ in \eqref{eq:time_dep_ggm}.  
A natural approach is to formulate a two-step process where one first estimates the parameters $\bbphi$, uses them to center the observations collected in $\bbX$, and then solves GL as in \eqref{eq:GL_problem}.  
A key observation is that any method capable of accurately estimating the mean (i.e., the parameters $\bbphi$) can be employed for this purpose \cite{curbelo2025adaptive, martino2021automatic}

In this work, we advocate for a more involved method that integrates Bayesian inference with frequentist estimation.
Let us assume some priors $g_{\bbphi}(\bbphi)$ and $g_{\bbTheta}(\bbTheta) \propto \exp(-\lambda \| \bbTheta \|_1)$, and a likelihood function $\ell({\bf X}|{\bm \Theta},\bbphi) = \prod_r\mathcal{N}({\bf x}_r|{\bf f}_r(\bbphi),\bbTheta^{-1})$, so that the posterior is given by $\pi(\bbphi,\bbTheta|{\bf X}) \propto \ell({\bf X}|{\bm \Theta},\bbphi) g_{\bbphi}(\bbphi)g_{\bbTheta}(\bbTheta)$.
Specifically, we aim to compute the maximum a posteriori (MAP) estimate of $\bbphi$ and the sparsity-regularized maximum likelihood estimator of $\bbTheta$, respectively given by
\begin{align}
    \phimap &= \argmax_{\bbphi} \log \pi(\bbphi | \bbX, \Thetaml), 
    \\
    \Thetaml &= \argmax_{\bbTheta} \log \ell (\bbX | \bbTheta, \phimap) - \lambda \| \bbTheta \|_1.
\end{align}
Note that the maximizer $\Thetaml$ coincides with the one in \eqref{eq:GL_problem} when we assume ${\bf f}_r(\bbphi)={\bf 0}$ for all $r$.
Here, $\pi(\bbphi | \bbX, \Thetaml)$ denotes the conditional posterior of $\bbphi$ and $\ell(\bbX | \bbTheta, \phimap)$ is the likelihood of $\bbTheta$ given $\phimap$.
Following~\cite{curbelo2025adaptive}, we estimate both $\phimap$ and $\Thetaml$ via an iterative algorithm, which we term \emph{Graphical Lasso with Adaptive Target Adaptive Importance Sampling (GL-ATAIS)}.  
GL-ATAIS is an alternating optimization algorithm where the following steps are sequentially performed for a maximum of $K$ iterations. 

\vspace{2mm}
\noindent\textbf{Step 1 - Mean estimation}.  
The first step is to estimate $\hbphi^{(k+1)}$ by maximizing the log-posterior $\pi(\bbphi | \bbX, \Thetaml^{(k)})$, where $\Thetaml^{(k)}$ is the estimate from the previous iteration.  
This amounts to solving the following optimization problem 
\begin{equation}\label{eq:phi_step}
  \min_{\bbphi} \left[ \! \frac{1}{2} \! \sum_{r=1}^R (\bbx_r \!-\! \bbf_r(\bbphi))^\top \Thetaml^{(k)} (\bbx_r \!-\! \bbf_r(\bbphi)) \!-\! \log g_{\bbphi}(\bbphi) \! \right], 
\end{equation}
where $\bbx_r$ are the columns of $\bbX$ and $g_{\bbphi} (\bbphi)$ represents a prior on the distribution of $\bbphi$.  

Since $\bbf_r(\bbphi)$ is not restricted to be convex or differentiable, we rely on an importance sampling (IS) procedure to find an approximate solution to \eqref{eq:phi_step}.  
More precisely, given a \textit{proposal} density $q(\bbphi | \tbmu^{(k)}, \tbSigma^{(k)})$ with mean $\tbmu^{(k)}$ and covariance $\tbSigma^{(k)}$, we draw $P$ samples $\{ \bbphi_k^{(p)} \}_{p=1}^P$ and compute $\hbphi^{(k+1)}$ as
\begin{equation}
    \hbphi^{(k+1)} = \bbphi_k^{(p_k^*)},~ \text{with}~p_k^*:= \argmax_{p\in\{1,...,P\}} \pi( \bbphi_k^{(p)} | \bbX, \Thetaml^{(k)}).
\end{equation}
The impact of the IS procedure becomes evident in the third step, where the samples $\{ \bbphi_k^{(p)} \}_{p=1}^P$ are used to compute the importance weights and update $\tbmu^{(k)}$ and $\tbSigma^{(k)}$.

\vspace{2mm}
\noindent\textbf{Step 2 - Precision estimation}.  
After estimating $\hbphi^{(k+1)}$, the mean of the GGM underlying our data $\bbX$ is given by $\bbf_r(\hbphi^{(k+1)})$.  
Thus, we center the observations by removing the estimated mean and compute the sample covariance as  
\begin{equation}
    \widehat{\bbSigma}^{(k+1)} = \frac{1}{R} \sum_{r=1}^R (\bbx_r - \bbf_r(\hbphi^{(k+1)})) (\bbx_r - \bbf_r(\hbphi^{(k+1)}))^{\top},
\end{equation}
and then estimate $\hbTheta^{(k+1)}$ via GL by solving  
\begin{equation}\label{eq:GL_centered}
    \hbTheta^{(k+1)} \! = \! \argmin_{\bbTheta \succeq 0} \frac{1}{2}\tr(\widehat{\bbSigma}^{(k+1)}\bbTheta) \!-\! \frac{1}{2} \! \log\det (\bbTheta) + \lambda \| \bbTheta \|_1.
\end{equation}
Clearly, this is equivalent to solving the original GL formulation in~\eqref{eq:GL_problem}, but replacing the sample covariance $\hbSigma_{\texttt{emp}}$ with its centered estimate $\widehat{\bbSigma}^{(k+1)}$.  
Since \eqref{eq:GL_centered} is strictly convex, a variety of efficient algorithms can be employed to obtain the global minimum~\cite{witten2011new,natali2022learning,navarro2025fair}.

\vspace{2mm}
\noindent\textbf{Step 3 - Adaptation}.
The final step updates the estimates $\phimap^{(k+1)}$ and $\Thetaml^{(k+1)}$, as well as the mean and covariance of the proposal density function $q( \bbphi | \tbmu^{(k)}, \tbSigma^{(k)})$.  
With the conditional log-posterior of $\bbphi$ given by
%
\begin{multline}
    \log \pi( \bbphi | \bbX, \bbTheta) = -\frac{1}{2}\sum_{r=1}^R (\bbx_r \!-\! \bbf_r(\bbphi))^\top \bbTheta (\bbx_r \!-\! \bbf_r(\bbphi)) \\
\quad + \frac{1}{2}\log\det (\bbTheta) \!+\! \log g_{\bbphi}(\bbphi) - \lambda \| \bbTheta \|_1,
\end{multline}
we update the estimates as follows.
We set
\begin{align}
    &\phimap^{(k+1)} = \hbphi^{(k+1)},
    &\Thetaml^{(k+1)} = \hbTheta^{(k+1)},
\end{align}
if $\log \pi( \hbphi^{(k+1)} \! | \bbX, \! \hbTheta^{(k+1)}) \! > \! \log \pi( \phimap^{(k)} | \bbX, \! \Thetaml^{(k)} \!)$.
Otherwise, we retain the previous estimates, i.e., $\phimap^{(k+1)} = \phimap^{(k)}$ and $\Thetaml^{(k+1)} = \Thetaml^{(k)}$.
\newline
Next, using the samples $\{ \bbphi_k^{(p)} \}_{p=1}^P$ from Step~1, we compute the importance weights
\begin{equation}
    w_k^{(p)} = \frac{\pi( \bbphi_k^{(p)} | \bbX, \hbTheta^{(k)})}{ q( \bbphi_k^{(p)} | \tbmu^{(k)}, \tbSigma^{(k)})}.
\end{equation}
The proposal distribution parameters are then updated as
\begin{align}
    \tbmu^{(k+1)} &= \phimap^{(k+1)}, \label{eq:prop_mean_update} \\
    \tbSigma^{(k+1)} &= \sum_{p=1}^P \bar{w}_k^{(p)} (\bbphi_k^{(p)} - \barbphi_k)(\bbphi_k^{(p)} - \barbphi_k)^\top + \delta_k\bbI, \label{eq:prop_cov_update}
\end{align}
where $\bar{w}_k^{(p)} = \frac{w_k^{(p)}}{\sum_{p=1}^P w_k^{(p)}}$, $\barbphi_k = \sum_{p=1}^P \bar{w}_k^{(p)} \bbphi_k^{(p)}$, and $\delta_k > 0$.   
Updating the proposal parameters following \eqref{eq:prop_mean_update} and \eqref{eq:prop_cov_update} improves the performance and robustness of IS, mitigating catastrophic scenarios and leading to more stable results, particularly as the problem dimension increases~\cite{el2018robust,llorente2023marginal}.  
Furthermore, updating $\phimap^{(k+1)}$ and $\Thetaml^{(k+1)}$ only when the evaluation of the regularized log-posterior improves helps avoid convergence to suboptimal stationary points. 
After running steps 1–3 for $K$ iterations, GL-ATAIS provides a frequentist estimate of the precision matrix $\Thetaml = \Thetaml^{(K)}$ and, consequently, the graph topology.
Moreover, the Bayesian estimation of $\phimap = \phimap^{(K)}$ not only yields a point estimate but also enables uncertainty quantification of the estimator.  
Developing a Bayesian inference method to learn $\bbTheta$ and assess the uncertainty of the estimated edges remains an intriguing direction for future research.

\subsection{Accelerating the convergence}
The convergence of the global optimization problem can be accelerated by implementing minor modifications in \eqref{eq:phi_step} in the first iterations.
Namely, in order to find a good region of the space for the initial point of the alternating optimization procedure, we can have a number of iterations $K_0<K$ of the algorithm considering
\begin{multline}
    \boldsymbol{\phi}_{\mathrm{MAP}}^{(k)} \!=\! \arg \min _{\boldsymbol{\phi}}\left[ \! \frac{1}{2} \! \sum_{r=1}^R\left\|\mathbf{x}_r-\mathbf{f}_r(\boldsymbol{\phi})\right\|^2-\log g_\phi(\boldsymbol{\phi}) \! \right],
\end{multline}
with $k=1, \ldots, K_0$, instead of finding the minimum of the optimization problem in \eqref{eq:phi_step}.
This is equivalent to set $\widehat{\boldsymbol{\Theta}}_{\mathrm{GL}}^{(k)}=\mathbf{I}_N$ for $k=0, \ldots, K_0-1$ in \eqref{eq:phi_step}, where $\mathbf{I}_N$ is the $N \times N$ identity matrix. Thus, in the first $K_0$ iterations, we focus only on finding a good point $\bbphi_{\text {MAP }}^{\left(K_0\right)}$.
Indeed, note that if there exists a point $\boldsymbol{\phi}^*$ such that $\sum_{r=1}^R\left\|\mathbf{x}_r-\mathbf{f}_r\left(\boldsymbol{\phi}^*\right)\right\|^2=0$, then this point $\boldsymbol{\phi}^*$ is also a root for $\sum_{r=1}^R\left(\mathbf{x}_r-\mathbf{f}_r\left(\boldsymbol{\phi}^*\right)\right)^{\top} \widehat{\boldsymbol{\Theta}}\left(\mathbf{x}_r-\mathbf{f}_r\left(\boldsymbol{\phi}^*\right)\right)=0$ for any possible precision matrix $\widehat{\boldsymbol{\Theta}}$.

\section{Numerical evaluation}
This section evaluates how the time-varying mean of the data impacts the performance of GL, illustrating the importance of estimating its mean and the benefits of GL-ATAIS.  

\vspace{2mm}
\noindent\textbf{Experiment setup.}
We generate Erd\H{o}s-Rényi (ER) graphs with $N = 10$ nodes and an edge probability of $p = 0.1$.  
The true precision matrix is then constructed as $\bbTheta = \bbA + \epsilon\bbI$, where $\epsilon > 0$ is a small constant ensuring that $\bbTheta$ is positive semidefinite.  
We sample $R$ graph signals from a Gaussian distribution $\ccalN(\bbf_r(\bbphi), \bbTheta^{-1})$, with $\bbphi \in \reals^4$ and a time-dependent mean function $\bbf_r(\bbphi)$ given by
\begin{equation}\label{equ:graph f}
\small
	\begin{array}{rcl}
		f_{r,[1]}(\bphi) &=& -\phi_4 \tau_r + 5 \phi_1^2 \\
		f_{r, [2]}(\bphi) &=& 2\phi_3 \sin(-\phi_2 \tau_r) \\
		f_{r,[3]}(\bphi) &=& \phi_1 - \phi_3 + \phi_1\cos(2 \tau_r) \\
		f_{r,[4]}(\bphi) &=& 3\phi_4 + 3\phi_2 + \phi_1 \exp(0.1 \tau_r) \\
		f_{r,[5]}(\bphi) &=& \phi_3^2 - 2 \phi_1 + 3\phi_2 - \exp(0.8 \phi_3) \exp(1 - \tau_r) \\
		f_{r,[6]}(\bphi) &=& 5 (\phi_4 + \phi_3) - \phi_2 \log(1 + 2 \tau_r) \\
		f_{r,[7]}(\bphi) &=& 3 \phi_2 - 0.2 \tau \sin(\phi_3) \\
		f_{r,[8]}(\bphi) &=& 3\phi_1 + 5 \phi_3 - 20\sin(\phi_4) \cos(2 \tau_r +\dfrac{\pi}{4}) \\
		f_{r,[9]}(\bphi) &=& \phi_2 + 4\phi_4 + 5 \exp\left (\dfrac{1}{1 + \phi_3}\right ) \tau_r \\
		f_{r,[10]}(\bphi) &=& 5\phi_1 + 10\phi_3 - 5\phi_4 \sin(\tau_r),
	\end{array}
\end{equation}
Here, $f_{r,[i]}$ denotes the value of the function $\bbf_r(\bbphi)$ at the node $i$, and $\tau_r$ specifies the timestamp of the $r$-th sample.
More precisely, the values of $\tau_r$ correspond to $R$ equally spaced timestamps in the interval $[0, 4]$.

We compare the following methods: i) Standard GL: the classical GL formulation \eqref{eq:GL_problem}, where the observed data is centered using the sample mean estimate; ii) GL with true mean: an oracle method where the true $\bbmu_r$ is used to center the observations before applying GL, hence serving as an upper bound for the performance of other methods; iii) GL-ATAIS: our proposed algorithm, which jointly estimates the mean and the precision matrix; and iv) ATAIS~\cite{curbelo2025adaptive}: a method that estimates $\bbTheta$ as the inverse of the inferred covariance matrix. 
Finally, we evaluate the performance using the average F-score of the estimated graph over 100 independent realizations.  

\begin{figure}
    \centering
    \includegraphics[width=0.7\linewidth]{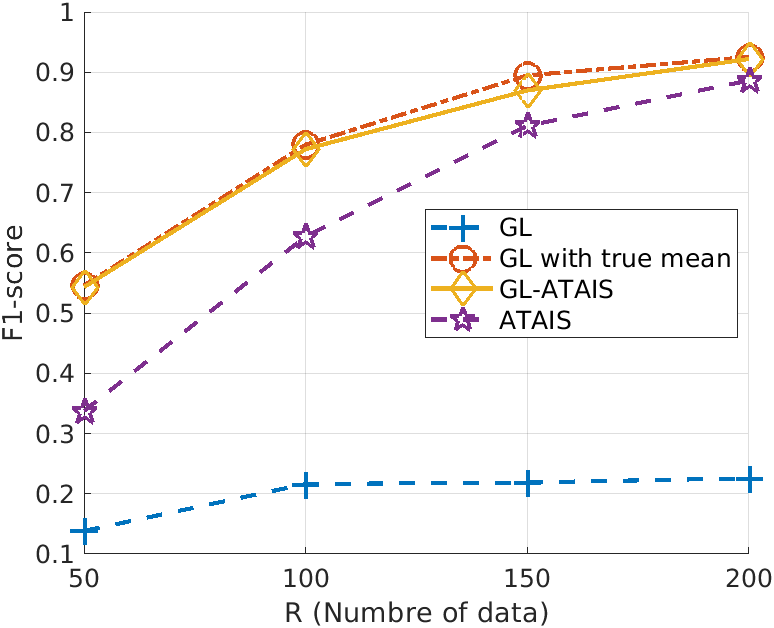}
    \caption{Average F1-score over 100 repetitions for four algorithms, when the number of observations varies in \{50, 100, 150, 200\}. The ATAIS repetitions are run for T=30 iterations, with N=30000 particles.}
    \label{fig:f1-score}
\end{figure}

\vspace{2mm}
\noindent\textbf{Test case 1 - Impact of the number of observations.}
In this experiment, we analyze how the number of observed signals $R$ influences the quality of the estimated graph.  
Fig.~\ref{fig:f1-score} depicts the F-score of different methods as $R$ increases from 50 to 200.  
As expected, performance improves with larger $R$, except for standard GL, which remains consistently poor.  
In contrast, GL with the true mean outperforms all alternatives, reinforcing our claim that standard GL cannot recover the graph topology (i.e., the support of $\bbTheta$), when the mean varies over time.
This result underscores the importance of properly estimating the mean before applying GL.  
While relying on the true mean is unrealistic in practical scenarios, it serves as an upper bound for graph recovery performance.  
Interestingly, GL-ATAIS achieves an F-score comparable to this oracle case, highlighting the effectiveness of our method.  
Moreover, GL-ATAIS significantly outperforms ATAIS, particularly when $R$ is small, demonstrating the advantages of using GL for precision matrix estimation instead of directly inverting the estimated covariance.

\begin{figure}
    \centering
    \includegraphics[width=0.75\linewidth]{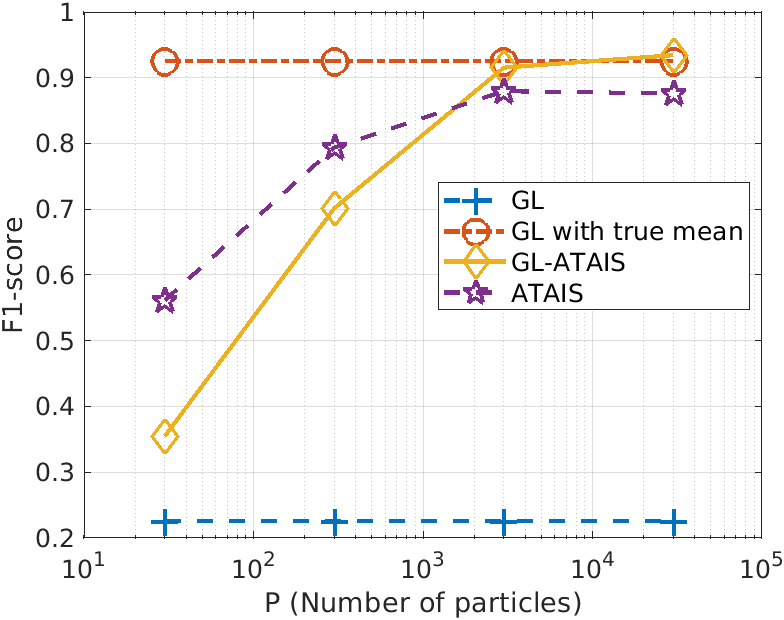}
    \caption{Average F1-score over 100 repetitions for four algorithms versus the number of particles in ATAIS.}
    \label{fig:relative-error}
\end{figure}

\vspace{2mm}
\noindent\textbf{Test case 2 - Impact of the number of particles.}
Next, we investigate how the number of particles $P$ used to estimate the mean parameters $\bbphi$ affects the recovery of $\bbTheta$.  
Fig.~\ref{fig:relative-error} presents the results, where $P$ ranges from 30 to 30000.  
Note that the number of particles only affects the performance of ATAIS and GL-ATAIS since they are the only methods based on IS.
Consistent with the previous experiment, we observe that standard GL fails to recover the graph when using the sample mean, while it performs best when the true mean is available.  
For large $P$, GL-ATAIS outperforms ATAIS and approaches the performance of GL with the true mean, validating the importance of jointly estimating the mean and precision matrix.  
However, when $P$ is small, the estimation of $\bbphi$ deteriorates, leading to a poor graph recovery.  
This highlights the trade-off between computational cost and estimation accuracy when selecting the number of particles in GL-ATAIS.

\section{Conclusion}
This work addressed the problem of learning graphs from nodal signals when such signals are drawn from a GGM with a time-varying mean.  
To tackle this challenge, we proposed GL-ATAIS, an iterative method that combines Bayesian inference with frequentist estimation to jointly learn the time-varying mean and the precision matrix.
Assuming that the mean is represented by a parametric function $\bbf_r(\bbphi)$, GL-ATAIS employs an importance sampling strategy to refine the estimate of the parameters $\bbphi$, leverages this estimate to center the observed data, and then applies GL to obtain the maximum likelihood estimator $\bbTheta$.  
Notably, this approach enables uncertainty quantification for the parameters $\bbphi$, and developing a Bayesian estimator for $\bbTheta$ remains an interesting direction for future research.  
Finally, our numerical experiments demonstrated that the zero-mean assumption of GL severely hinders its ability to recover the true graph topology, while GL-ATAIS effectively mitigates the bias introduced by time-varying means, resulting in significantly improved graph recovery.

\bibliographystyle{IEEEbib}
\bibliography{myIEEEabrv,biblio}

\end{document}